\documentclass[sigconf]{acmart}

\usepackage[font={small,it}]{caption}
\usepackage{graphicx}
\usepackage{comment}
\usepackage[utf8]{inputenc} 
\usepackage[T1]{fontenc}    
\usepackage{hyperref}       
\usepackage{url}            
\usepackage{amsfonts}       
\usepackage{nicefrac}       
\usepackage{microtype}      
\usepackage{amsmath}
\usepackage{subcaption}
\usepackage{algorithm}
\usepackage{algpseudocode}
\usepackage{geometry}

\setcopyright{rightsretained}

\copyrightyear{2024}
\acmYear{2024}
\setcopyright{rightsretained}
\acmConference[ICVGIP 2024]{Indian Conference on Computer Vision Graphics and Image Processing}{December 13--15, 2024}{Bengaluru, India}
\acmBooktitle{Indian Conference on Computer Vision Graphics and Image Processing (ICVGIP 2024), December 13--15, 2024, Bengaluru, India}
\acmPrice{}
\acmDOI{10.1145/3702250.3702287}
\acmISBN{979-8-4007-1075-9/24/12}

\begin{document}
\title{ Revised Regularization for Efficient Continual Learning through Correlation-Based Parameter Update in Bayesian Neural Networks}
\titlenote{Produces the permission block, and
  copyright information}


\author{Sanchar Palit}
\affiliation{%
  \institution{IIT Bombay}
  \city{Mumbai}
  \country{India}
  }
  \orcid{0009-0002-3007-9330}
\email{204070004@iitb.ac.in}

\author{Biplab Banerjee}
\affiliation{%
  \institution{IIT Bombay}
  \city{Mumbai}
  \country{India}
  }
\email{bbanerjee@iitb.ac.in}

\author{Subhasis Chaudhuri}
\affiliation{%
  \institution{IIT Bombay}
  \city{Mumbai}
  \country{India}
  }
\email{sc@ee.iitb.ac.in}

\renewcommand{\shortauthors}{}

\begin{abstract}
We propose a Bayesian neural network-based continual learning algorithm using Variational Inference, aiming to overcome several drawbacks of existing methods. Specifically, in continual learning scenarios, storing network parameters at each step to retain knowledge poses challenges. This is compounded by the crucial need to mitigate catastrophic forgetting, particularly given the limited access to past datasets, which complicates maintaining correspondence between network parameters and datasets across all sessions. Current methods using Variational Inference with KL divergence risk catastrophic forgetting during uncertain node updates and coupled disruptions in certain nodes. To address these challenges, we propose the following strategies. To reduce the storage of the dense layer parameters, we propose a parameter distribution learning method that significantly reduces the storage requirements. In the continual learning framework employing variational inference, our study introduces a \textit{regularization term} that specifically targets the dynamics and population of the \textit{mean and variance} of the parameters. This term aims to retain the benefits of KL divergence while addressing related challenges. To ensure proper correspondence between network parameters and the data, our method introduces an \textit{importance-weighted Evidence Lower Bound} term to capture data and parameter correlations. This enables storage of common and distinctive parameter hyperspace bases. The proposed method partitions the parameter space into \textit{common and distinctive subspaces}, with conditions for effective backward and forward knowledge transfer, elucidating the network-parameter dataset correspondence. The experimental results demonstrate the effectiveness of our method across diverse datasets and various combinations of sequential datasets, yielding superior performance compared to existing approaches.
\end{abstract}

%
%



\maketitle

\section{Introduction}


Humans possess the cognitive ability to learn new concepts while retaining their previously acquired knowledge. Furthermore, the integration of novel knowledge can often result in an increased ability to understand and solve problems. Conversely, machine learning models have a limited capacity to retain previously learned problem-solving skills when trained to address new problems. In addition, these models lack the capability to enhance their problem-solving proficiency while learning to solve new tasks. Preserving knowledge from previous tasks while training on subsequent ones has remained a significant challenge, prompting extensive attention from the research community.

In order to tackle the issues of Catastrophic Forgetting \cite{kirkpatrick2017overcoming}, and \cite{li2017learning}  promote Continual Learning, various approaches have been proposed in the literature. Regularization-based model \cite{kirkpatrick2017overcoming,serra2018overcoming} aims to preserve the previously learned knowledge by penalizing any modification of important parameters corresponding to previously learned tasks. Dynamic architecture-based models \cite{yoon2017lifelong,rusu2016progressive,schwarz2018progress} bifurcates different parameters of the models to different tasks. Additionally, memory-based models \cite{wu2019large,lee2019overcoming,castro2018end} have been introduced that mitigate catastrophic forgetting by storing some raw samples \cite{lopez2017gradient} of the previous tasks for rehearsal or using generative models to synthesize data from the past while training the current parameters. These approaches ensure that the knowledge of the past is maintained while learning new tasks.


Gaussian parameterized Bayesian techniques (\cite{blundell2015weight,tseran2018natural,blei2017variational}) avoid drawbacks associated with point-based methods by incorporating uncertainty in both parameters and network intermediates. Parameters are represented as samples drawn from distributions defined by mean and variance, effectively addressing parameter uncertainty. These techniques utilize evidence lower bound (ELBO) loss of variational inference, where the KL divergence term acts as a natural regularizer. The other component of ELBO, log-evidence, establishes a connection between the data and the model. Variational Inference is commonly studied in the context of two general Gaussian distributions.

However, utilizing KL as a regularizer necessitates the surrogate and the approximate to align within the same support. This KL divergence between successive Gaussian posteriors restrains the movement of mean and variance in certain directions, aiding the continual learning framework in retaining knowledge and updating mean and uncertainty judiciously. Nonetheless, the gradient flow becomes ill-posed due to the high-dimensional manifold. To ensure smoothness in trajectory, \cite{muller1997integral} proposed different lower bound estimators for the low-dimensional manifold. Similarly, Integral Probability Metrics \cite{arbel2019maximum} and Maximum Mean Discrepancy \cite{glaser2021kale} introduced different gradient flows using only samples of the target distribution. We emphasize the motion of the mean during gradient computation, introducing flexibility to its movement alongside retaining a certain level of uncertainty to enhance knowledge acquisition.

Subsequently, several pertinent questions arise, prompting further discussion: 
\textit{Can we feasibly implement a revamped regularization concerning both mean and variance by modifying existing regularization, such as KL divergence, to overcome its drawbacks while retaining its advantages?} Additionally, in variational inference, the log-evidence term aims to identify the parameter values that best fit the dataset. However, in a continual learning scenario, access to past data is restricted. \textit{Can we devise a mechanism to achieve optimal alignment between the network parameters and the data across all sessions?} Furthermore, obtaining backward knowledge transfer in the traditional continual learning framework poses a significant challenge.

Optimization-based methods \cite{blei2017variational} within Bayesian techniques exhibit superior performance when compared to sampling-based approaches\cite{brooks1998markov}. Among Optimization-based methods, Variational Inference \cite{hoffman2013stochastic} is widely used. It approximates the prior and posterior distribution by minimizing the distance between the true and the approximate distribution using an ensemble of known distributions. However, Bayesian neural networks employing variational inference necessitate twice the number of parameters compared to point-based regular neural networks. The introduction of additional uncertainty allows for flexibility in parameter learnability, as discussed in \cite{ebrahimi2019uncertainty}. An approach to managing this uncertainty involves preserving highly uncertain parameters for ongoing learning, while keeping certain parameters fixed, as they have already acquired crucial attributes for the current task. However, this strategy introduces several drawbacks, particularly the challenge of controlling parameter learnability. Highly certain parameters may compel the updating of connected uncertain parameters during backpropagation as the parameters undergo updates. 

We propose utilizing the mean and variances of the parameter distribution to characterize the importance and uncertainty of parameters in the Continual Learning process with Bayesian neural network. The proposed approach involves carefully selecting the mean (position) and variance (spread) of the parameter distribution to accurately capture the uncertainty of the parameters. Subsequently, during experimentation, we utilize visual illustrations to depict the progression of the optimal posterior parameter distribution of each task and from this we can deduce the global optimal parameter distribution for all tasks.\footnote{Session and tasks are interchangeably used to denote different time steps in continual learning.} It can be intuitively understood that the more spread out (high variance) a parameter distribution is, the more commonality it shares with other session parameters, thus enabling the possibility of achieving commonality in the future to enhance parameter learnability. Hence, the number of uncertain parameters in our method is systematically decreased, guaranteeing an equitable level of learnability across all tasks.

Various variants of variational inference are present in the literature, among which Stochastic Variational Inference (SVI) \cite{hoffman2013stochastic} is noteworthy. In the domain of large datasets, SVI optimizes model parameters through the application of stochastic gradient descent. However, for effective backpropagation, a reliable gradient estimate of the evidence lower bound with respect to the variational parameters is essential. Black Box Variational Inference (BBVI) \cite{ranganath2014black} emerges as a solution with minimal constraints on the form of the variational posterior, earning its designation as black box variational inference due to its flexibility. In our proposed method, we introduce an efficient approach by replacing the KL divergence term with revised regularization terms. This substitution facilitates gradient computation and imposes fewer constraints on the variational posterior within the context of continual learning.

\textbf{Contributions:}
Our work has devised a Continual Learning framework with distinctive attributes, outlined as follows: a) In the domain of continual learning with bayesian neural network, the work introduce an enhanced regularization term aimed at maintaining the beneficial aspects of KL divergence while mitigating associated challenges. This regularization term facilitates controlled adjustments of mean parameters and ensures sparsity in variance values. Additionally, it constrains the increment of variance values to achieve uniform uncertainty parameter distribution across different tasks. b) The proposed approach introduce a method to store dense layer parameters efficiently, reducing storage complexity by incorporating a compact network capable of learning the previous posterior weight distribution. This enables proficient generation of samples from past distributions. For convolutional neural networks, our method utilizes a representation matrix to optimize session-specific layer parameters. c) An importance-weighted Evidence Lower Bound term is introduced, facilitating the capture of correlations between data and network parameters. The parameter subspace is partitioned into common and distinctive subspaces, with common bases contributing to overall performance across sessions, and distinctive bases providing insights into specific session characteristics.
   

\section{Related work:}
\textbf{Continual Learning:}
From a high-level view, Continual Learning models fall into memory-based, regularization-based, and gradient-based methods. Memory-augmented continual learning involves storing past data or features to enhance training, as seen in methods like Gradient Episodic Memory (GEM) \cite{lopez2017gradient} and Averaged-GEM (A-GEM) \cite{chaudhry2018efficient}. Rehearsal-based models such as Experience-Replay \cite{rolnick2019experience} and Meta-Experience-Replay \cite{riemer2018learning} address forgetting by training on both previous and new data, requiring supplementary storage and adding to the overall cost. 
Regularization-based approaches penalize significant changes in important parameters from previous steps. Elastic Weight Consolidation \cite{kirkpatrick2017overcoming} and Synaptic Intelligence \cite{zenke2017continual} identify and assign importance scores to vital parameters. Gradient Projection Memory \cite{saha2021gradient} and Orthogonal Gradient Descent \cite{farajtabar2020orthogonal} leverage gradient steps aligned with important subspaces, stored in memory. Orthogonal Weight Modulation (OWM) \cite{zhang2021understanding} modifies weights in directions orthogonal to past task inputs.
Knowledge transfer is explored through Continual Learning with Backward Knowledge Transfer (CUBER) \cite{lin2022beyond}, allowing backward transfer if gradient steps align with previous task subspaces. Uncertainty-Guided Continual Learning \cite{ebrahimi2019uncertainty} adjusts learning rates based on parameter uncertainty for improved backward transfer. In these methods, preserving previous network parameters is crucial. However, efficiently storing network parameters to reduce costs or establish universally beneficial knowledge across sessions remains a challenge. We aim to leverage the repetitive structure of parameter drifts to reduce storage costs and facilitate knowledge sharing across sessions.


\textbf{Bayesian approaches:}
Research on integrating Bayesian techniques \cite{mackay1992bayesian,snoek2012practical} into neural networks has been active for several decades. Several notable approaches in this field include Variational Inference with Gaussian Processes \cite{nguyen2014automated}, Probabilistic Backpropagation \cite{hernandez2015probabilistic}, and Bayes by Backprop \cite{blundell2015weight}. A method based on Variational Inference, known as Variational Continual Learning \cite{nguyen2017variational,ahn2019uncertainty}, obtains the posterior distribution by multiplying the prior distribution with the data likelihood of the current session during training. Variational continual learning derives the current posterior by multiplying the previous posterior with the likelihood of the dataset. VCL incorporates supplementary coresets of the raw dataset, which are updated using the previous coreset and the current dataset. This approach updates the distribution with non-coreset datapoints, adding the challenge of managing coreset datapoints. Another variant proposed in \cite{tseran2018natural} employs an online update of the mean and variance using Variational Online Gauss-Newton \cite{khan2018fast} update. Additionally, \cite{chen2019facilitating} utilizes the \cite{liu2016stein} gradient to generate data samples for a known distribution within VCL. In our approach, we leverage Variational Inference with a unique adaptation in the KL divergence term, resulting in a significant performance improvement compared to these methods.

\section{Background:{Variational Inference}}
In this section, we review the variational inference using a Bayesian neural network. In this framework, we use a Bayesian model $\mathbb{P}(\mathcal{Y}|\mathcal{X}, \mathcal{W})$ where we use a neural network model with weights $\mathcal{W}$ for the input-output combination of the dataset $\mathcal{D}= (\mathcal{X}, \mathcal{Y})$. We assume that the weights are drawn from a prior distribution $\mathcal{W} \sim \mathbb{P}(\mathcal{W}|\beta)$, parameterized by some $\beta$. During training obtaining the posterior using Baye's rule will lead to intractability. So Variational Inference obtains the approximate posterior with the following objective,
\begin{equation}
\mathcal{Q}(\mathcal{W}|\theta) = \arg \min_{\theta} \mathrm{KL}(\mathcal{Q}(\mathcal{W}|\theta)||\mathbb{P}(\mathcal{W}|\mathcal{D})) 
\end{equation}
or equivalently by minimizing the following loss function as shown in \ref{sec:kullback_leibler},
\begin{equation}
\mathcal{L}_{\mathrm{ELBO}}(\theta, \mathcal{D}) = -\mathbb{E}_{\mathcal{Q}(\mathcal{W}|\theta)}[\log \mathbb{P}(\mathcal{D}|\mathcal{W})] + \mathrm{KL}(\mathcal{Q}(\mathcal{W}|\theta)||\mathbb{P}(\mathcal{W}|\mathcal{D}))\label{eq:elbo_main}
\end{equation}
Here we take $\mathcal{Q}(\mathcal{W}|\theta)$ as Gaussian parameterized by $\theta = (\mu, \sigma)$\footnote{ $\theta = (\mu, \sigma)$ or $\theta = (\mu, \rho)$ where $\sigma = \ln (1+ \exp(\rho))$}, inferred from Bayes by Backprop (BBB) \cite{blundell2015weight}, offering a known probability distribution over network parameters. Backpropagation on these parameters demonstrates satisfactory performance.

\subsection{\textsc{Continual Learning with Variational Inference}:}
Consider a continual learning framework where data arrives sequentially, denoted as $\mathcal{D}=\{(\mathcal{X}_i, \mathcal{Y}_i)\}_{i=1}^{\mathcal{N}}$. Here, $\mathcal{X}_i =\{\mathcal{X}_{i,j}\}_{j=1}^{\mathcal{N}_i}$ represents input vectors, and $\mathcal{Y}_i =\{\mathcal{Y}_{i,j}\}_{j=1}^{\mathcal{N}_i}$ represents corresponding labels at session $i$.
At time-step $t$, our objective is to learn the posterior $\mathbb{P}(\mathcal{W}_t|\theta, \mathcal{D}_t)$ after sequentially being trained on $\mathcal{D}_1, \mathcal{D}_2, \cdots \mathcal{D}_{t-1}$ where $\mathcal{D}_t = \{\mathcal{X}_{t, j},\mathcal{Y}_{t,j}\}_{j=1}^{\mathcal{N}_t}$. Hence the posterior needs to cumulatively acquire knowledge of $\Tilde{\mathcal{D}}_t = \bigcup_{i=1}^{t}\mathcal{D}_i$. 
To update the parameter distribution at time step $t$, we lack access to $\Tilde{\mathcal{D}}_{t-1}$.
To address the computational challenge of obtaining the posterior at each step, we assume $\mathcal{W}_t$ is sampled from an approximate posterior $\mathcal{Q}(\mathcal{W}_t|\theta_t) \approx \mathbb{P}(\mathcal{W}_t| \Tilde{\mathcal{D}_t})$. Our objective is to minimize the following loss function:
\begin{align}
 \mathcal{L}_{\mathrm{ELBO}} (\theta_t, \mathcal{D}_t) = \mathbb{E}_{\mathcal{Q}(\mathcal{W}|\theta_t)}&[-\log \mathbb{P}(\mathcal{D}_t|\mathcal{W}_t)] + \nonumber\\
 &\mathrm{KL} [\mathcal{Q}(\mathcal{W}_t|\theta_t)||(\mathcal{Q}(\mathcal{W}_{t-1}|\theta_{t-1})] \label{eq:elbo_continual}
\end{align}
Here, Our model weight $\mathcal{W}_i$ gets updated at each time-step such that $\mathcal{W}_t \sim \mathcal{Q}(\mathcal{W}_t|\theta_t)$. 
\label{problem_definition}

\section{Efficiently learning prior weights in parameter space}
 The existing Bayesian methodology for computing the likelihood term in Equation \ref{eq:elbo_continual} relies on Monte-Carlo sampling. However, this approach is highly sensitive to the sample size, governed by the law of large numbers, and a small sample size can lead to inaccurate estimates. This dependence on sample size not only compromises accuracy but also escalates both space and time complexity, posing challenges to computational efficiency. Furthermore, the network, in this Bayesian framework, is compelled to store the prior distribution $\mathcal{Q}(\mathcal{W}_{t-1}|\theta_{t-1})$ using $\theta_{t-1} = (\mu_{t-1}, \sigma_{t-1})$. Consequently, the number of parameters to be maintained becomes precisely twice the size of the network weights. 
We leverage a Bayesian neural network wherein the parameter $\theta$ governing $\mathcal{Q}(\mathcal{W}|\theta)$, where $\theta = (\mu, \sigma)$, is learned through stochastic variational inference \cite{hoffman2013stochastic} and back-propagation. This approach allows for the derivation of a more tightly bounded lower limit on the variational objective, showcasing accelerated performance in contrast to other sampling-based Bayesian techniques, such as Markov Chain Monte Carlo \cite{blei2017variational}. To comprehensively assess the effectiveness of our approach we carefully examine two models: a fully connected neural network and a convolutional neural network. These models are known for their effectiveness in computer vision tasks, and we're exploring how they perform differently in Bayesian learning because of their distinct architectures.
\subsection{Efficient Parameter Update for Fully Connected Neural Networks:} 
Let's consider a fully connected neural network model within the Bayesian framework, operating on a dataset $\mathcal{D}$ characterized by an input vector {$\mathcal{X} \in \mathbb{R}^{m}$} and a label vector {$\mathcal{Y} \in \mathbb{R}^{n}$}. Specifically, $\theta = \{\mu, \sigma\}$ consists of $\{\mu, \sigma\} = (\{\mathcal{W}_{\mu}, \mathcal{B}_{\mu}\},\{\mathcal{W}_{\sigma}, \mathcal{B}_{\sigma}\})$ of weights and biases, where ($\mathcal{W}_{\mu}, \mathcal{W}_{\sigma}, \mathcal{B}_{\mu}, \mathcal{B}_{\sigma} \in \mathbb{R}^\tau $ {where $ \tau = n \times k \cdots \{h \: \mathrm{times}\}\times m$}, with $h$ being the length of the hidden layer and $k$ being the length of each hidden layer). We decompose each parameter matrix into three parts, such as {$\mathcal{W}_{\mathrm{in}} \in \mathbb{R}^{k \times m}$} representing shared weights on the input side, $\mathcal{W}_{h} \in \mathbb{R}^{k \times k}$ representing shared weights in the hidden layer and {$\mathcal{W}_{c} \in \mathbb{R}^{n \times k}$} representing final non-shared classification weights. A novel distribution learning network is employed to obtain $\mathcal{W}_{\mu}$ for different connections for $\mathcal{W}_{\mathrm{in}}$ and $\mathcal{W}_{h}$. $\mathcal{W}_{\mu}$ of $\mathcal{W}_{\mathrm{in}}$ exhibits two connections: one on the input side of length $m$ and the other on the output side of length $k$. Each of the $k$ output nodes receives input from $m$ incoming nodes on the input side. Every $k$-dimensional tensor with size $m$ shares a consistent trajectory in parameter space. 
This is attributed to the fact that all outgoing nodes observe the same set of input nodes with varying weight connections.
\begin{figure*}
\centering
\begin{subfigure}{0.35\textwidth}
  \centering
  \subfloat[]
  {\includegraphics[width=6cm]{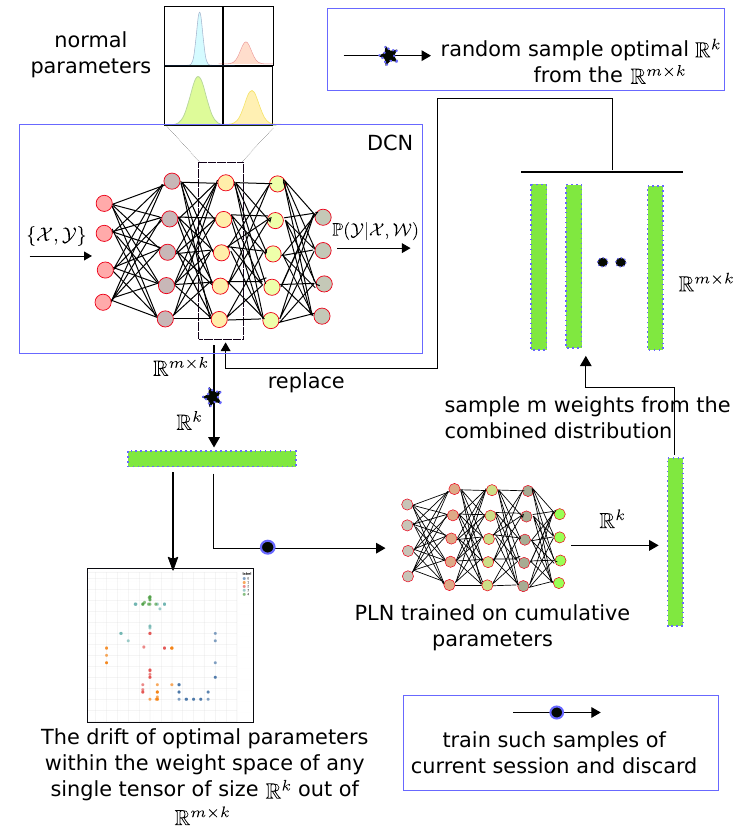}
  \label{fig:cont_learn_full}}
  \end{subfigure}
  \begin{subfigure}{0.55\textwidth}
  \subfloat[]{\includegraphics[width=1.0\textwidth]{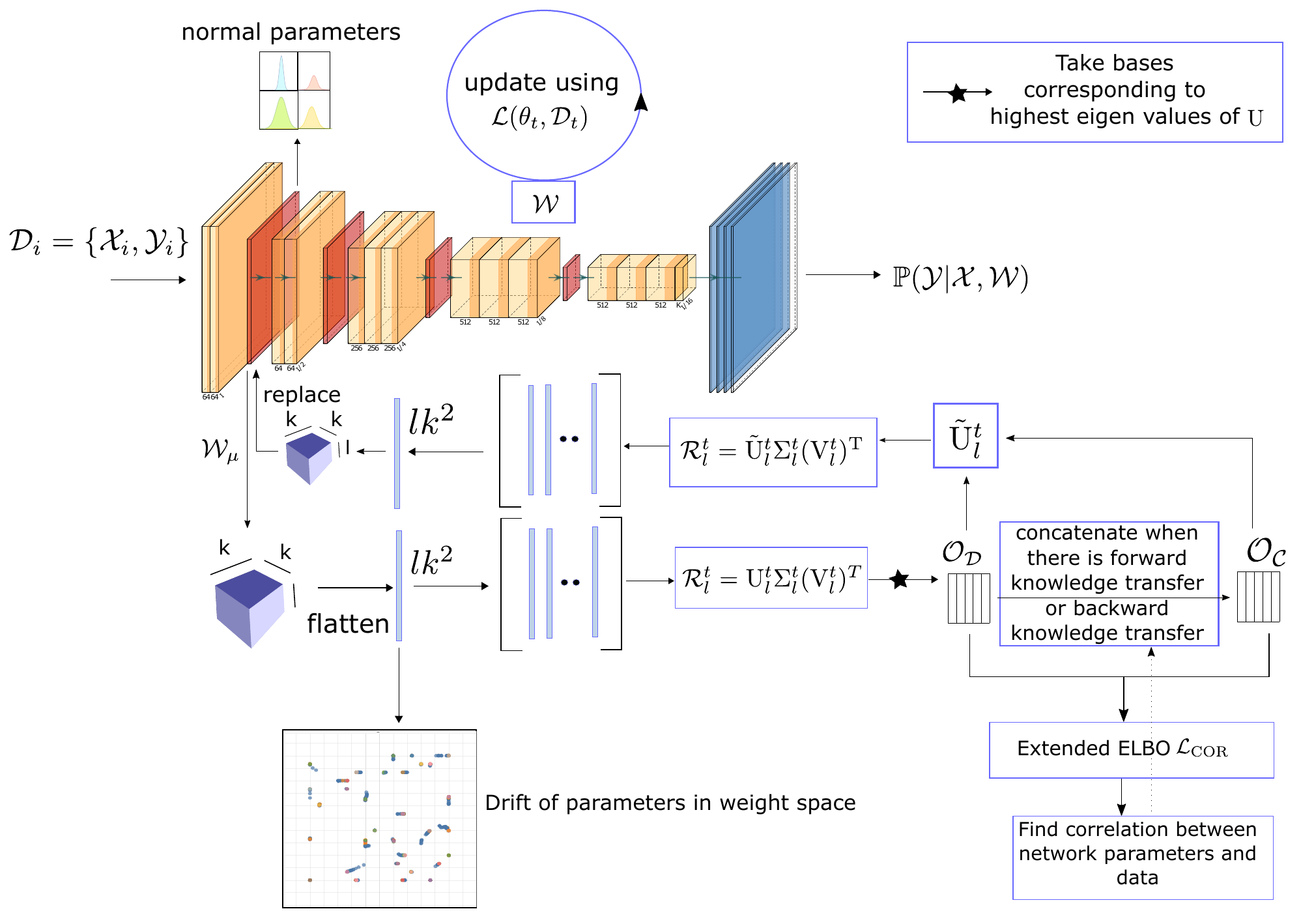}
  \label{fig:continual_learning_convolutional_main}}
\end{subfigure}
\caption{\footnotesize{(a) Efficient parameter updating for a convolutional neural network involves updating the model parameters with the revised loss function. (b) Efficient parameter updating for a convolutional neural network involves updating the model parameters with the revised loss function. Following this, the basis for the differentiated subspace is determined using SVD on the representation matrix. Upon establishing correspondence, this basis is then shifted to the common subspace.
}}
\label{fig:cont_learn_full_convolutional_main}
\end{figure*}
Consequently, any perturbation on the incoming side equally affects all outgoing nodes. Therefore, during backpropagation, all connections need to be updated similarly. To address this, we propose utilizing a smaller-sized network to generate tensors of size {$\mathbb{R}^{m \times k}$(or $\mathbb{R}^{k \times m}$), i.e., $m$ tensors, each of size $k$(or $k$ tensors, each of size $m$)}. This network employs a distribution generator capable of learning the distribution of any $k$-dimensional tensor. The distribution learning network comprehensively captures the trajectory of all $k$-dimensional tensors across all sessions in the continual learning framework, enabling it to generate tensors that are optimal across all sessions. As all $m$ vectors of dimension $k$ move identically in the parameter space, the same network can be employed to generate all $m$ vectors consistently. This is depicted in Fig.~\ref{fig:cont_learn_full}. The identical procedure is iteratively applied to the connections $\mathcal{W}_{h}$. 
We use two different networks for the aforementioned problem. First, we use a Deep Classification Network (DCN) for which weights are sampled from $\mathcal{W} \sim \mathcal{Q}(\mathcal{W})$ parameterized by $\theta = \{\mu, \sigma\}$ which is mapping the input sample to the output label $\mathcal{X} \rightarrow \mathcal{Y}$. Simultaneously, we learn another function $\mathcal{G}_{\psi}(\mathbb{R}^{k} \rightarrow \mathbb{R}^{k})$, parameterized by $\psi$, called a Parameter Learning Network (PLN) which can map any uniformly selected random sample $(\mathbb{R}^{k})$ to the parameter space $(\mathbb{R}^k)$. Here we should note that the PLN can generate the parameter of the DCN in such a way that the size of the PLN is far less than the size of the parameter layer i.e. the overall parameter size of PLN $\ll \mathbb{R}^k$. Thus, the method effectively stores the PLN, which comprises significantly fewer parameters compared to other Bayesian-based techniques. By composing the two functions we get $f_{\mathcal{W},\theta}(\mathcal{X}) =\mathcal{G}_{\psi}\{\mathcal{Q}(\mathcal{W}|\theta)\}(\mathcal{X})$, which represent our entire model for the Continual Learning Classification tasks.
\subsection{Efficient Parameter Update for Convolutional Neural Networks:}
Let us consider a Convolutional Neural Network with the input tensor $\mathcal{X} \in \mathbb{R}^{(C_{i} \times h_i \times w_i)}$ and filters $\mathcal{W}_{\mu} \in \mathbb{R}^{ m \times(l \times k \times k)}$. Here $k$ is the kernel size and $l$ is $C_i$ for the initial convolutional layer and $m$ and $l$ increases as we go deeper into the convoutional layer. After passing it through different convolutional layers it is finally fed to fully connected layers before feeding it to the final application-specific classification head. We try to focus on the drifts of the initial convolutional layers as next dense layers can be used similar to what explained in the previous section. For $\mathcal{W}_{\mu}$ of each convolutional layer we have a different representation matrix. $\mathcal{W}_{\mu} \in \mathbb{R}^{ m \times (l \times k \times k)}$ is reshaped to $\mathcal{W}_{\mu} \in \mathbb{R}^{ m \times lk^2}$ generate $m$ one dimensional tensor to be stored in the representation matrix. The representation matrices are explained in the subsequent section. The parameter drift in the parameter space for the convolutional layers is different than the layers of a dense layer. The flattened convolutional masks shows overlapping pattern in the parameter space. Hence in this scenario we focus on the trajectory of the convolutional layers. {The process is illustrated in Figure~\ref{fig:continual_learning_convolutional_main} and the components are discussed in the subsequent section.}

\textit{Optimal parameter trajectory in parameter hyperspace:}
To determine the optimal parameters of $\mathcal{Q}(\mathcal{W}|\theta)$, the objective is to obtain the parameters $\mathcal{W}^{(i)} \sim \mathcal{Q}(\mathcal{W}^{(i)})$, where $\mathcal{W}^{(i)} \in \mathbb{R}^{m}$, such that the loss is minimized. This entails analyzing the trajectories of $\mathcal{W}^{(i)}$ across all previous sessions $0, 1, \cdots, t-1$, focusing specifically on the parameter paths that lead to the minimum loss during training at each task or towards the conclusion of training. These paths, located in proximity to the global minima of the loss function, are retained and revisited subsequently to obtain optimal parameter values suitable for all sessions.
In this context, we define $\mathcal{T}_{j, k} = \{ \mathcal{W}_{j,1}, \mathcal{W}_{j,2}, \cdots, \mathcal{W}_{j,k}\}$ a random trajectory of length $k$ sampled from the parameter space at the $j$-th session when the model is trained on dataset $\mathcal{D}_j$. Finally, $\mathbb{P}(\mathcal{T}_{j,k}|\mathcal{D}_j)$ gives a distribution overall trajectories of length $k$ that can be sampled from the parameter space when the model is trained on the full dataset $\mathcal{D}$. Our objective is the find the trajectory $\mathcal{T}_{\mathrm{opt}}$ which is in minimum distance from all the previously seen trajectories $\mathcal{T}_{1,k}, \mathcal{T}_{2,k}, \cdots, \mathcal{T}_{(i-1), k}$ and the current trajectory $\mathcal{T}_i$  or the overlapping portion of the trajectory $\mathcal{T}_{\mathrm{opt}} = \underset{j}{\arg \min} \bigcap_{i \in \mathrm{tasks}}  \texttt{distance}(\mathcal{T}_{j}- \mathcal{T}_{i, k}) $. The distance metric indicates the similarity of trajectories, aiming to minimize the objective function.
We specifically select the optimal parameters at the conclusion of epochs to fulfill this objective. 

\begin{figure*}[!h]
\centering
\begin{subfigure}{0.32\textwidth}
  \subfloat[]
  {\includegraphics[width=5.5cm]{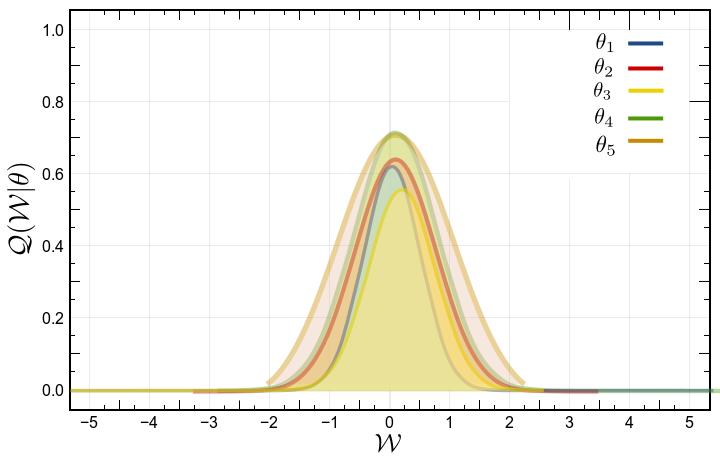}
  \label{fig:normal_begining}}
\end{subfigure}
\begin{subfigure}{0.32\textwidth}
  \subfloat[]
  {\includegraphics[width=5.5cm]{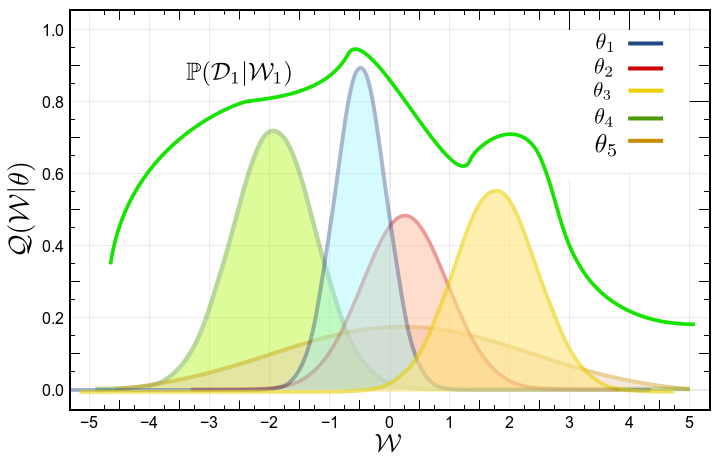}
  \label{fig:normal_training_1}}
\end{subfigure}
\begin{subfigure}{0.32\textwidth}
  \subfloat[]
  {\includegraphics[width=5.5cm]{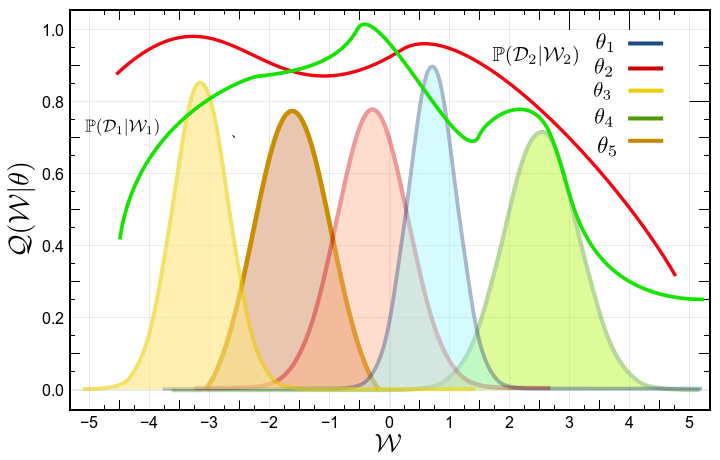}
  \label{fig:normal_training_2}}
\end{subfigure}
\caption{\footnotesize{Illustration of backward knowledge transfer. (a) parameters at the beginning (b) parameters after training on task 1 (c) parameters after training on task 2. The parameters that exhibited uncertainty after task 1 and were subsequently learned during task 2 contribute to the model's ability to approach a closer alignment between the log evidence curve and its surrogate.}}
\label{fig:normal_beg_1_2}
\end{figure*}

\textbf{Gradient Analysis of the KL divergence term and Regularization Proposals:}
The KL divergence term $\mathrm{KL}(\mathcal{Q}(\mathcal{W}_t|\theta_t)||\mathcal{Q}(\mathcal{W}_{t-1}|\theta_{t-1}))$ is a measure of information loss when posterior $\mathcal{Q}(\mathcal{W}_t|\theta_t)$ is used instead of prior $\mathcal{Q}(\mathcal{W}_{t-1}|\theta_{t-1})$. The gradient with respect to $\theta_t = (\mu_t, \sigma_t)$ is given as:
\begin{align*}
\begin{bmatrix} 
\scalebox{0.95}{$\frac{\partial \mathrm{KL}}{\partial \mu_t}$} \\ \scalebox{0.8}{$\frac{\partial \mathrm{KL}}{\partial \sigma_t}$} \end{bmatrix} = \begin{bmatrix} \scalebox{0.8}{$\sigma_t^{-1}(\mu_t - \mu_{t-1})$} \\ \scalebox{0.9}{$-\frac{1}{2}\left(\sigma_t^{-1} - \sigma_t^{-1}(\mu_t - \mu_{t-1})(\mu_t - \mu_{t-1})^T\sigma_t^{-1}\right)$} \end{bmatrix}
\end{align*}
\textit{Proposed mean regularization:}
The gradient signifies that an increase in the value of $\mu_t$ will proportionally increase the KL divergence between $ \mathcal{Q}(\mathcal{W}_t|\theta_t)$ and $\mathcal{Q}(\mathcal{W}_{t-1} |\theta_{t-1})$ and vice-versa. The scale factor ${\sigma_t}^{-1}$ denotes the extent to which changes in $\mu_t$ influence the KL divergence. If $\sigma_t$ is small, indicating low uncertainty in the posterior, variations in $\mu_t$ will exert a more pronounced impact on the KL divergence and vice-versa. In this context, challenges arise when one layer is regarded as certain while the subsequent layer is deemed uncertain. Updating the uncertain node may lead to the loss of information in both incoming and outgoing nodes, potentially resulting in instances of catastrophic forgetting. To mitigate this, additional regularization is introduced, focusing on the magnitude of the mean change. This supplementary regularization offers regularization to nodes regardless of their uncertainty status. Consequently, we introduce two regularization terms: one penalizing the magnitude of the mean difference and the other penalizing the magnitude of the ratio of the mean difference to the previous sigma. These regularization terms are mathematically formulated as presented in Equation~\ref{eq:up_loss}. 

\textit{Proposed variance regularization:}
The gradient of the KL divergence with respect to $\sigma_t$ involves second-order terms, specifically $(\mu_t-\mu_{t-1})(\mu_t-\mu_{t-1})^{T}$, representing the products of mean differences along both dimensions. These terms reflect the covariance or correlation between alterations in $\mu_t$ and $\mu_{t-1}$. The inclusion of these second-order terms implies that the influence of mean differences is contingent on the covariance between changes in $\mu_t$ and $\mu_{t-1}$. Positive covariance, indicating concurrent changes, strengthens the impact on the KL divergence, suggesting that the model may readily adapt to alterations when $\mu_t$ and $\mu_{t-1}$ exhibit positive correlation. The limitations associated with this term include the fact that an increased variance $\sigma_t$ reduces the responsiveness of the KL divergence to alterations in $\mu_t$ and $\mu_{t-1}$. Furthermore, updating the variance as the inverse of the variance may lead to many parameters becoming more certain over time. Conversely, if we intend to preserve some degree of learnability in parameters, a gradual increase in certainty for certain parameters would be desirable.
We add two extra regularization terms: one for the sparsity of the previous variance ($\lVert \sigma_{t-1} \rVert_1$) and the other for the sparsity of the ratio of variance change relative to the previous variance ($\lVert \frac{\sigma_{t} - \sigma_{t-1}}{\sigma_{t-1}} \rVert_1$). Introducing sparsity regularization on the previous variance encourages adaptability for future parameter learning in lifelong learning scenarios. The second term restricts significant changes in variance values, aiming to minimize variations between steps. Essentially, adding this term to the loss function encourages stability or restrained changes in variance values between consecutive steps, which can be advantageous in scenarios where we aim to manage the rate of change or avoid substantial fluctuations in model parameter uncertainty.
Therefore, the updated loss function with the added mean and variance regularization terms is expressed as:
\begin{align}
 \mathcal{L}(\theta_t,&\mathcal{D}_t) = \mathbb{E}[-\log \mathbb{P}(\mathcal{D}_t|\mathcal{W}_t)] + \nonumber \\ & \underset{\text{revised} \: \mu- \: \text{regularization}}{\underline{\lVert \mu_t - \mu_{t-1} \rVert_2^2 + \lVert \frac{(\mu_t - \mu_{t-1})^2}{{\sigma_t^2}} \rVert_2^2}} + \underset{\text{revised} \: \sigma- \: \text{regularization}}{\underline{\lVert \sigma_{t-1} \rVert_1 + \lVert \frac{\sigma_{t} - \sigma_{t-1}}{\sigma_{t-1}} \rVert_1}}
 \label{eq:up_loss}
\end{align}
It's worth noting that in {equation \ref{eq:elbo_continual}}, the loss function is referred to as the Evidence Lower Bound, as it consistently remains smaller than the log evidence term, given that the KL divergence is always greater than zero. Our proposed regularizer in equation \ref{eq:up_loss} confirms that the revised terms are invariably positive, thus ensuring that the loss function remains a lower bound of the log evidence.

\textbf{Extended Distribution for Constructing Importance-Weighted Log-Evidence and ELBO:}
We utilize importance sampling within the subspaces defined by the common and distinctive subspaces to construct the prior distribution from the preceding session. This enables us to obtain the ELBO and log-evidence with this distribution, as detailed in \cite{domke2018importance} and \cite{sobolev2019importance}. The log-evidence term aims to identify the parameter values that most suitably fit the dataset. By employing importance weighting with the log-evidence, we can confirm the optimal alignment between the network parameters and the data across all sessions. 
The log-evidence is approximated using, $\mathcal{R}_{t,\mathrm{M}} = \frac{1}{\mathrm{M}}\sum_{i=1}^{\mathrm{M}} \frac{\mathcal{Q}(\mathcal{W}^{t-1}_{i}, \mathcal{D}_{t-1})}{\mathcal{Q}(\mathcal{W}^{t}_{i})}$. This employs importance-weighted approximation of the random variable $\mathcal{R}_t = \frac{\mathcal{P}(\mathcal{D}_t, \mathcal{W}_t)}{\mathcal{Q}(\mathcal{W}_t)}$ and a tighter lower bound for Jensen's inequality is achieved as the resulting distribution is more centered towards the mean.
\begin{align}
 \mathcal{L}_{\mathrm{COR}}(\mathcal{W}^t,& \mathcal{D}_{t-1}) = \mathbb{E}_{\mathcal{Q}(\hat{\mathcal{W}}_{1:\mathrm{M}})}[\log \mathcal{R}_{t,\mathrm{M}}] + \mathrm{KL}[\mathcal{Q}(\mathcal{W}^{t}_{1})|| \\
 & \mathcal{Q}(\mathcal{W}^{t-1}_{1}|\mathcal{D}_{t-1})] \nonumber +\mathrm{KL}[\mathcal{Q}(\mathcal{W}^{t}_{2:\mathrm{M}}|\mathcal{W}^t_1)||\mathcal{Q}(\mathcal{W}^{t}_{2:\mathrm{M}})]
\end{align}
This function is utilized to quantify the correlation between the model parameters and the dataset, while Eq.~\ref{eq:up_loss} is employed for the model parameter updates.

\textbf{Knowledge Transfer and Sample Correspondence:}

\textit{Robust correspondence with previous samples:}
 This enhancement facilitates robust backward knowledge transfer, emphasizing a strong correlation between the current weight parameters and their predecessors. Consequently, we leverage the bases of the common subspace, eliminating the need for additional parameters in the current session, as we can efficiently reuse the bases from the preceding session. Mean updates are performed based on a stringent correlation check, employing the following equation:
\begin{align}
 \rho (\mathcal{L}(\mathrm{Proj}_{\mathcal{O}_{\mathcal{C}}^l}(\mathcal{W}^{t}_l), \mathcal{D}_{t}),\mathcal{L}_{\mathrm{COR}}(\mathcal{W}^{t}_l, \mathcal{D}_{t-1}) ) \geq \epsilon, \nonumber \\ 
 \mu_l^t \leftarrow \mu_l^t -\lambda\nabla_{\mu} \mathcal{L}(\mu_l^t)-\alpha \nabla_{\mu} \mathcal{L}(\mathcal{P}_l^t\mu_l^t)
 \label{eq:rob_corr}
\end{align}
In this context, $\epsilon$ functions as the correlation threshold. 
In this scenario, updating the loss using the bases of the common subspace is feasible. However, this approach may result in a suboptimal approximation for the current dataset in subsequent iterations, as unique information about the dataset remains undiscovered. To address this limitation, SVD is applied to the residual representation matrix $\Bar{\mathcal{R}}_l^t = \mathcal{R}_l^t\mathcal{O}_{\mathcal{C}}^l(\mathcal{O}_{\mathcal{C}}^l)^T$ to extract novel information, and these novel bases are stored in the differentiated subspace. 
It is also pertinent to observe the layerwise correlation among the weight parameters. In the preceding discussion, we updated the parameters of a specific layer while maintaining the parameters of other layers fixed. Notably, the layer-wise correlation is independent of the dataset. Formally, it can be expressed as:
$\rho( \mathcal{L}(\mathcal{W}_{i, \mathcal{C}}^{t}, \mathcal{D}_{t}) , \mathcal{L}(\mathcal{W}_{j, \mathcal{C}}^{t}, {\mathcal{D}}_{t})) = \texttt{const}$ $\forall i, j=1,\cdots, n$.

\textit{Backward knowledge transfer:}
Backward Knowledge Transfer characterizes the enhancement of problem-solving capabilities for earlier tasks through the assimilation of crucial information acquired while addressing the current task. This observation suggests an enhancement in performance on the preceding dataset when incorporating bases from the differentiated subspace, which encompasses the newly acquired bases from the current session. This is illustrated in Fig.~\ref{fig:normal_beg_1_2}.
\begin{equation}
 \mathcal{L}(\mathrm{Proj}_{\mathcal{O}_{\mathcal{C}}^l\oplus\mathcal{O}_{\mathcal{D}}^l}(\mathcal{W}_{l}^{t}), \mathcal{S}_{t-1}^{j}) \leq \mathcal{L}(\mathrm{Proj}_{\mathcal{O}_{\mathcal{C}}^l}(\mathcal{W}_{l}^{t}), \mathcal{S}_{t-1}^{j})
 \label{eq:back_knw_tra}
\end{equation}
It is important to highlight that, in order to facilitate backward knowledge transfer, a distinct coreset is preserved, akin to the approach in VCL \cite{nguyen2017variational}. Here, $\mathcal{S}_{t-1}^{j}$ represents the datapoints from session $j \leq t-1$. Subsequently, the acquired bases in the differentiated subspace are concatenated with the common subspace, as they prove beneficial for at least two session parameters. In this context, the concept of confidence is employed, wherein the update of the parameter is confined within the established confidence interval. Specifically, weight updating is permissible in any direction within this confidence interval. This approach enhances the performance of both the current session and the preceding session while mitigating the risk of catastrophic forgetting.
\section{\textsc{Experimental Results:}}
\subsection{\textsc{Experimental setup:}}
\textbf{Datasets:}
Our experimental configuration is designed to emulate a class incremental learning paradigm, where each session involves a fixed number of classes from a designated dataset. To enhance our understanding, we extend our investigation to a multi-task learning framework, with each session centered around a distinct dataset. We conduct our experiments on two variations of the MNIST dataset, namely Permuted MNIST and split MNIST. Additionally, we explore three distinct variants of the CIFAR dataset, including CIFAR-100, split CIFAR 10-100, alternating CIFAR 10/100, and an 8-mixture dataset. Notably, no data augmentation techniques are employed in our experimental protocol. 

\textbf{Baseline:}   
We compare our results with established Bayesian Continual Learning methodologies, as well as non-Bayesian methods, both necessitating the retention of prior parameters.
For non-Bayesian methods, our comparison includes HAT \cite{serra2018overcoming} and Elastic Weight Consolidation (EWC) \cite{kirkpatrick2017overcoming} (although EWC can be regarded as Bayesian-inspired). Additionally, we compare against state-of-the-art methods such as Synaptic Intelligence (SI) \cite{zenke2017continual}, Incremental Moment Matching (IMM) \cite{lee2019overcoming}, Learning without Forgetting (LWF) \cite{li2017learning}, Less Forgetting Learning \cite{jung2016less}, Progressive Neural Network (PNN) \cite{rusu2016progressive}, and PathNet \cite{fernando2017pathnet} for CNN-based experiments.
Within the Bayesian framework, we evaluate Uncertainty-based Continual Learning with Adaptive Regularization (UCL) \cite{ahn2019uncertainty}, as well as Variational Continual Learning (VCL) \cite{nguyen2017variational}, and two VCL variants, Gaussian natural gradients (VCL-GNG) \cite{tseran2018natural} and Variational Adam (VCL-Vadam) \cite{tseran2018natural}. Additionally, we compare against the latest work within this framework, namely, Uncertainty-guided Continual Learning with Bayesian Neural Network (UCB) \cite{ebrahimi2019uncertainty}.
In this context, we conduct comparisons with ST-BCNN, ST-BFNN, ST-CNN, ST-FNN, which represent Bayesian CNN \cite{shridhar2019comprehensive}, Bayesian FNN, and conventional point-based fully connected neural network and convolutional neural network, respectively. These models are trained on new data without any mechanisms in place to mitigate catastrophic forgetting. Additionally, we compare against the combined training of these variants, denoted as CT-BCNN, CT-BFNN, CT-CNN, CT-FNN. These models are trained on the full dataset in a single-step supervised learning fashion, serving as an upper bound as they do not experience forgetting due to their training setup.

\textbf{Performance measurement metrics:}
After continual learning, algorithm performance is evaluated using two key metrics: average accuracy (ACC) and backward transfer (BWT), calculated empirically as follows:
\begin{equation}
     \scalebox{0.9}{\textsc{Acc}} = \frac{1}{\mathrm{T}}\sum_{j=1}^{\mathrm{T}}\mathcal{A}_{\mathrm{T},j} \; \text{, and}\; 
     \scalebox{0.9}{\textsc{BWT}} = \frac{1}{\mathrm{T}-1}\sum_{i=1}^{\mathrm{T}-1} \mathcal{A}_{\mathrm{T}, i}- \mathcal{A}_{i,i}
     \label{eq:acc_back_tra}
\end{equation}
Here, $\mathcal{A}_{i,j}$ represents the test accuracy on task $j$ after the model has completed task $i$. The model is trained on all previous tasks after completing the current task. Consequently, we construct a lower triangular matrix $\mathcal{A} \in \mathbb{R}^{\mathrm{T}\times \mathrm{T}}$, where each row corresponds to a timestep, and each column corresponds to a task number. The accuracies for each session are obtained by averaging the rows of the matrix $\mathcal{A}$. BWT quantifies the impact of the learner on previous tasks after being trained on the current task. A positive BWT indicates an improvement in performance on previous tasks, while a negative value suggests catastrophic forgetting. BWT computes the average shift from each diagonal element (representing the accuracy of the model when it initially encounters each task) in the current accuracy matrix.

\begin{table*}
    \begin{subtable}[H]{0.24\textwidth}
        \centering
        \scalebox{0.8}{
        \begin{tabular}{|c|c|c|}
        \hline
        \textbf{Method}&\textbf{BWT}$\uparrow$&\textbf{ACC} $\uparrow$\\
        \hline
             VCL-GNG & -& 90.50\\ 
             VCL-Vadam & -& 86.34 \\ 
             VCL & -7.90& 88.8\\ 
             IMM & -7.14 & 90.51 \\ 
             EWC & -&88.2 \\ 
             HAT & 0.03& 97.3 \\ 
             UCL & -& 94.5 \\ 
             UCB &0.03&97.42 \\ 
             LWF &-31.17 & 65.65 \\ 
             ST-BFNN & -0.58& 90.01 \\
             \textbf{ECL-RR}& \textbf{0.03} & \textbf{97.8}\\
         \hline
         CT-BFNN & 0.00 & 98.12 \\
         \hline
        \end{tabular}}
        \caption{}
        \label{tab:table_acc_perm}
    \end{subtable}
\hfill
    \begin{subtable}[H]{0.20\textwidth}
        \centering
        \scalebox{0.8}{
        \begin{tabular}{|c|c|c|}
        \hline
        \textbf{Method}&\textbf{BWT}$\uparrow$&\textbf{ACC}$\uparrow$\\
        \hline
             VCL-GNG&- &96.5\\
             VCL-Vadam& -&99.17 \\
             VCL& -0.56& 98.2\\
             IMM& -11.2 & 88.54 \\
             EWC& -4.20&95.78 \\
             HAT& 0.00& 99.59 \\
             UCL& -& 99.7 \\
             UCB&0.00&99.63 \\
             ST-FNN&-9.18 & 90.6 \\
             ST-BFNN & -6.45 & 93.42\\
             \textbf{ECL-RR}& \textbf{0.00} & \textbf{99.7}\\
         \hline
         CT-BFNN & 0.00 & 99.88 \\
         \hline
        \end{tabular}}
        \caption{}
        \label{tab:table_acc_split}
    \end{subtable}
\hfill
    \begin{subtable}[H]{0.19\textwidth}
        \centering
        \scalebox{0.8}{
        \begin{tabular}{|c|c|c|}
        \hline
        \textbf{Method}&\textbf{BWT}$\uparrow$&\textbf{ACC}$\uparrow$\\
        \hline
             LWF& -37.9 & 42.93\\
             PathNet & 0.00& 28.94\\ 
             LFL &-24.22 & 47.67\\ 
             IMM& -12.23& 69.37\\
             HAT& -0.04& 78.32\\
             PNN & 0.00& 70.73\\ 
             EWC& -1.53&72.46\\
             ST-BCNN& -7.43&68.89 \\
             UCB &-0.72&79.44 \\ 
             \textbf{ECL-RR} & \textbf{-0.03} & \textbf{80.16}\\
             \hline
             CT-BCNN & 1.52 & 83.93\\
             \hline
        \end{tabular}}
        \caption{}
        \label{tab:table_acc_cifar_alternating}
    \end{subtable}
\hfill
    \begin{subtable}[H]{0.18\textwidth}
        \centering
        \scalebox{0.8}{
        \begin{tabular}{|c|c|c|}
        \hline
        \textbf{Method}&\textbf{BWT}$\uparrow$&\textbf{ACC}$\uparrow$\\
        \hline
             LFL&-10.0& 8.61\\
             PathNet& 0.00&20.22\\
             IMM& -38.5 &43.93\\
             LWF& -54.3&28.22\\
             HAT& -0.14&81.59\\
             PNN& 0.00&76.78 \\
             EWC& -18.04&50.68 \\
             ST-BCNN& -23.1&43.09 \\
             UCB& -0.84&  84.04\\
             \textbf{ECL-RR} & \textbf{0.24} & \textbf{84.58}\\
            \hline
            CT-BCNN & 0.82& 84.85\\
        \hline
        \end{tabular}}
        \caption{}
        \label{tab:table_acc_mixture}
    \end{subtable}
\caption{average accuracy at the final session of Continual learning on the datasets (a)Permuted MNIST, (b)Split MNIST, (c)alternating CIFAR 10/100, (d)8-mixture dataset}
\label{tab:table_acc_perm_split_alt_mix}
\end{table*}

\begin{figure*}
\centering
\begin{subfigure}{0.325\textwidth}
  \subfloat[]
  {\includegraphics[width=5cm,height=5cm]{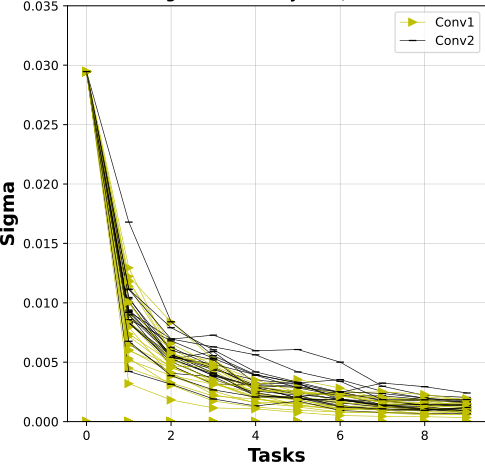}
  \label{fig:sigma_12}}
\hfill
\end{subfigure}
\begin{subfigure}{0.325\textwidth}
  \centering
  \subfloat[]
  {\includegraphics[width=5cm,height=5cm]{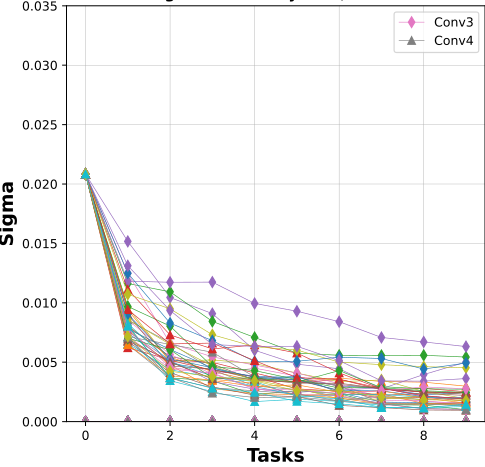}
  \label{fig:sigma_34}}
\hfill
\end{subfigure}
\begin{subfigure}{0.325\textwidth}
  \centering
  \subfloat[]
  {\includegraphics[width=5cm,height=5cm]{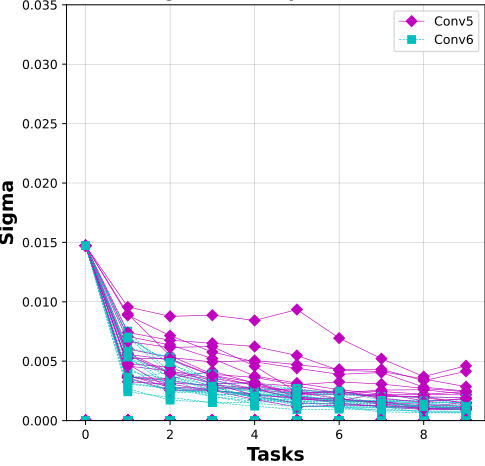}
  \label{fig:sigma_56}}
\end{subfigure}
\caption{Sigma at different convolutional layers for CIFAR100. (a) layers 1,2 (b) layers 3,4 (c) layers 5,6. {Best viewed when zoomed in.}} 
\label{fig:sigma_diff_layer}
\end{figure*}

\subsection{Main result:}
\textit{Permuted MNIST:}
Table~\ref{tab:table_acc_perm} presents the ACC and BWT metrics for ECL-RR, benchmarked against state-of-the-art models.
After undergoing sequential training across 10 tasks, EWC demonstrate minimal performance distinctions, achieving 90.2\%. SI achieves a slightly lower accuracy at 86.0\%, as reported by \cite{ahn2019uncertainty}. HAT attains a slightly better performance (ACC=91.6\%), and UCL achieves a decent accuracy of 94.5\%. Despite VCL with a coreset showing a 2\% improvement, ECL-RR outperforms all other baselines with an accuracy of 97.8\%. 
ECL-RR outperforms other Bayesian methods in accuracy with minimal forgetting (BWT=0.03\%), notably surpassing VCL (ACC=88.80\%) with BWT=-7.9\%. UCB (ACC=91.44\%) initially achieves parity with ECL-RR but experiences a decline in performance after task 6, underscoring the constraints of implementing UCB in a single-headed network. When compared to VCL-Vadam (ACC=86.34\%), VCL-GNG (ACC=90.50\%), and UCL (ACC=94.5\%), and ECL-RR consistently excels. With 1.9M parameters, \cite{ebrahimi2019uncertainty} reports 97.42\% for UCB and 97.34\% for HAT, while ECL-RR achieves 97.8\%, close to CT-BFNN (ACC=98.1\%), which serves as an upper bound. Importantly, incorporating memory in our experiments for backward knowledge transfer evaluation enhances performance. ST-BFNN, not fully penalized against forgetting, shows reasonable negative BWT, outperforming IMM and LWF.

\textit{split MNIST:}
Table~\ref{tab:table_acc_split} presents results for Bayesian and non-Bayesian neural networks, including simple training (ST-BFNN, ST-FNN) and combined training (CT-BFNN, CT-FNN). Despite the stability of MNIST, Bayesian methods consistently outperform point-based methods, highlighting their effectiveness. ECL-RR achieves an average accuracy of 99.7\% over the 5 tasks, matching UCL and slightly outperforming UCB and HAT with accuracies of 99.63\% and 99.59\%, respectively. Significantly surpassing EWC and VCL, ECL-RR outperforms all baselines while achieving zero forgetting, akin to HAT. Among VCL variants, VCL-Vadam (ACC=99.17\%) outperforms the original VCL (ACC=98.20\%) and VCL-GNG (ACC=96.50\%).

\textit{Alternating CIFAR 10/100:}
Table~\ref{tab:table_acc_cifar_alternating} displays ECL-RR's ACC and BWT outcomes on the alternating CIFAR-10/100 dataset compared to various continual learning baselines. PNN and PathNet exhibit reduced forgetting, albeit at the cost of limited learning in the ongoing session. PathNet's constrained task budget during model initialization impedes learning distinct attributes in each session. Notably, LWF and LFL show limited improvement after training in the initial two sessions on both CIFAR-10 and CIFAR-100, as the initial feature extractor does not exhibit substantial improvement in subsequent sessions. IMM achieves nearly the same performance as ST-BCNN, suggesting that the additional mechanism for reducing forgetting does not provide significant improvement. EWC and PNN perform slightly better than ST-BCNN, while HAT demonstrates noteworthy improvement with minimal forgetting. HAT achieves commendable performance without explicitly remembering past experiences, yet it is surpassed by our ECL-RR (ACC=80.16\%), which achieves an ACC almost 1.84\% higher than HAT. UCB exhibits marginal forgetting but attains an accuracy of 79.4\%, with ECL-RR surpassing it by a margin of 0.8\%. Thus, we can conclude that for highly intricate and complex computer vision datasets, especially those involving deeper CNN architectures, ECL-RR demonstrates commendable performance.

\textit{8-mixture dataset:} 
We present the results for continual learning on the 8-mixture dataset in Table~\ref{tab:table_acc_mixture}. Utilizing the performance of ST-BCNN (43.1\%) as a lower bound reference, we observe that EWC and IMM achieve results close to this lower bound. On the other hand, LFL and LWF exhibit a significant drop in accuracy, indicating their challenges in handling difficult datasets with pronounced catastrophic forgetting. PNN, HAT, and UCB are the only baselines that effectively mitigate catastrophic forgetting in this dataset. ECL-RR once again outperforms HAT by 3\% and UCB by 0.6\%.

\textbf{Visualization and Analysis of Parameter Drift in Bayesian Neural Networks:} 
We utilize Generative Topographic Mapping (GTM) \cite{bishop1998gtm} to examine the \textsc{Drift} of means across different layers for Bayesian-CNN and Bayesian-FNN, providing visual insights into the model parameters. The means $\{\mu_l^t\}_{l=1}^{\mathrm{L}}$ are stored for each layer across all sessions for visualization purposes.
In the case of Bayesian-FNN, we observe that parameters remain confined within a compact region for each session. Notably, the means of all nodes exhibit a similar drift pattern at each layer, even when considering the transpose of different layers.
Further analysis indicates that as the number of tasks increases, the drift of the means diminishes, signifying confinement within a smaller region. This consistent drift pattern is empirically demonstrated for all 3072 nodes in the input layer and 256 nodes in the hidden layer.
Conversely, in Bayesian-CNN, parameter values corresponding to different convolutional layers show substantial overlap with parameter values from previous sessions.
The consequence is that as more sessions are included in training, the optimal parameters tend to choose a previous value with high probability. 
This emphasizes the importance of distinguishing between the differentiated and common subspaces for effective continual learning.

\textbf{Uncertainty Analysis: Standard Deviations Across Convolutional Layers:}
To capture the essence of uncertainty, we present Fig.~\ref{fig:sigma_diff_layer}, illustrating the standard deviations for different convolutional layers across various sessions for the CIFAR-100 dataset. We initialize the standard deviations as 0.03, 0.02, and 0.015 for layers 1-2, 3-4, and 5-6, respectively. Notably, for layers 1 and 2, the values drop below 0.015 after session 1, while the reduction in other layers is more distributed, ranging between 30\% to 60\%. Subsequently, some layers exhibit an increase in variance values for certain nodes in the deeper layers.
The significant decrease in uncertainty in the initial layer after the first session suggests increased certainty, given the repetitive features encountered at this layer. On the other hand, the fluctuating nature of uncertainty in deeper layers indicates the possibility of having certain nodes with increased uncertainty without compromising the model's ability to learn distinctive attributes. This fluctuation adds extra flexibility by allowing some attributes to be forgotten, making room for new knowledge. Moving to linear layers and the output layer, where precision is crucial for predictions, the {gradual decrease} in uncertainty is desired to ensure that any alteration does not adversely impact previous predictions.
\section{Conclusion:}
In conclusion, this study introduces Efficient Continual Learning in Bayesian neural network, leveraging the advantages of Variational Inference. While acknowledging the challenge of a large parameter count, we have successfully minimized this issue. Moreover, we propose a modified regularization method that effectively addresses limitations in existing KL divergence regularization, controlling mean and variance dynamics. Our approach not only ensures stability and faster execution but also outperforms current state-of-the-art methods by a considerable margin.

\bibliographystyle{ACM-Reference-Format}
\bibliography{ICVGIP-Latex-Template}

\end{document}